\documentclass[lettersize,journal]{IEEEtran}
\usepackage[english]{babel}
\usepackage[T1]{fontenc}    
\usepackage{cite}
\usepackage[bookmarks=false,draft]{hyperref} 
\usepackage{amsfonts}       
\usepackage[pdftex]{graphicx}
\usepackage{amsthm,array,booktabs,tabularx,caption}
\newcolumntype{M}[1]{>{\centering\arraybackslash}m{#1}}
\usepackage{makecell}

\usepackage[cmintegrals]{newtxmath}
\usepackage{bm} 
\usepackage{paralist}

\usepackage{algorithm}
\usepackage{algorithmic}
\usepackage{stfloats}
\usepackage{xcolor}
\usepackage{soul}
\usepackage{etoolbox}
\usepackage{url}
\usepackage{paralist}
\usepackage{enumitem}
\usepackage{subcaption} 
\def\BibTeX{{\rm B\kern-.05em{\sc i\kern-.025em b}\kern-.08em
    T\kern-.1667em\lower.7ex\hbox{E}\kern-.125emX}}

\theoremstyle{definition}

\theoremstyle{definition}

\begin{document}
\title{Multi-Objective Optimization Using Adaptive Distributed Reinforcement Learning}

\author{Jing~Tan, Ramin~Khalili and~Holger~Karl
\thanks{Manuscript accepted by IEEE Transactions on Intelligent Transportation Systems; date of current version 15 December 2023.}
\thanks{J. Tan and R. Khalili are with the Huawei Technology Munich Research Center in Germany.}
\thanks{H. Karl is with the Hasso Plattner Institute, University of Potsdam in Germany.}
}

\maketitle

\begin{abstract}
The Intelligent Transportation System (ITS) environment is known to be dynamic and distributed, where participants (vehicle users, operators, etc.) have multiple, changing and possibly conflicting objectives. Although Reinforcement Learning (RL) algorithms are commonly applied to optimize ITS applications such as resource management and offloading, most RL algorithms focus on single objectives. In many situations, converting a multi-objective problem into a single-objective one is impossible, intractable or insufficient, making such RL algorithms inapplicable. We propose a multi-objective, multi-agent reinforcement learning (MARL) algorithm with high learning efficiency and low computational requirements, which automatically triggers adaptive few-shot learning in a dynamic, distributed and noisy environment with sparse and delayed reward. We test our algorithm in an ITS environment with edge cloud computing. Empirical results show that the algorithm is quick to adapt to new environments and performs better in all individual and system metrics compared to the state-of-the-art benchmark. Our algorithm also addresses various practical concerns with its modularized and asynchronous online training method. In addition to the cloud simulation, we test our algorithm on a single-board computer and show that it can make inference in 6 milliseconds.
\end{abstract}
\begin{IEEEkeywords}
V2X, Distributed Systems, Reinforcement Learning, Multi-Objective
\end{IEEEkeywords}

\section{Introduction}
\label{sec:intro}

An intelligent transportation system (ITS) comprises vehicle users, mobile users, edge and cloud service providers~\cite{arthurs2021taxonomy}. They all have individual objectives and private, changing preferences, and they act selfishly to achieve their objectives by competing for limited communication and computation resources in the network~\cite{xiong2020intelligent}. With advances in communication and autonomous technologies, ITS is connecting more people and devices, and centrally optimizing resource allocation through service providers is no longer practical.

Such an environment is well suited to multi-agent systems (MAS). They are distributed in nature, use agents to represent individual interests, and model complex interaction between players. MAS reduces model complexity and data requirements by breaking down a centralized problem into local, individual problems. Such systems are naturally compatible with game-theoretic approaches that have similar assumptions such as player independence, selfishness, and limited information-sharing \cite{bowling2000analysis}. 

Due to the complexity and dynamicity of ITS, agents in the MAS need to learn from and adapt to the environment. The study of multi-agent reinforcement learning (MARL) algorithms for resource allocation decisions in ITS is gaining traction~\cite{haydari2020deep}. RL algorithms are known for their ability to learn sequential tasks without supervision and purely based on feedback from the environment~\cite{sutton2018reinforcement}. MARL makes learning strategies more effective, especially in distributed environments~\cite{althamary2019survey}. 
But most such algorithms assume that all players have a single objective, although many real-world problems are multi-objective in nature~\cite{cho2017survey}. Converting a multi-objective problem into a single-objective one \begin{inparaenum}[1)] \item is impossible when utility or user preference over each objective is unknown \emph{a priori}, changing fast or incommensurate, or \item is intractable with high dimensionality or non-convexity, or \item performs worse because a single-objective learning algorithm cannot track the development of reward on multiple objectives~\cite{ansari2018vector,hayes2022practical}. \end{inparaenum} Few studies focus on solving multi-objective problems in ITS, and most of them use scalarization methods to simplify into a single-objective problem~\cite{khamis2014adaptive,aziz2018learning,pandey2020deep}. 
On the other hand, some multi-objective learning algorithms such as \cite{parisi2014policy} require learning and storing many models and extensive retraining whenever user preferences change; some others such as \cite{pirotta2015multi} are computationally expensive. Model-agnostic meta-learning (MAML) \cite{finn2017model} addresses these challenges---it trains a single model that is adaptable to different new tasks with few-shot learning. However, these algorithms are often applied to single-agents; they are rarely studied in a decentralized, multi-agent and non-stationary environment characteristic of ITS. 

In our study, we see a centralized approach as an unrealistic option for a real-life ITS system design. Hence, we focus our effort on a more practical approach, using independent multiagents with partial information. This approach limits information sharing and reduces communication overhead. More specifically, we use a parallel and distributed stochastic gradient ascent method [16] for training using Federated Learning framework. This is a logically centralized training method implemented in a distributed manner. In deployment, we execute the algorithm distributedly (i.e., each vehicle user infers its decisions independently), to avoid transferring data from / to a centralized entity that increases overhead and latency.  A decentralized approach as ours also avoids complexity in modeling and solving a centralized optimization problem from the operator side.

With our multi-objective design, vehicle users in an ITS environment can choose and weigh their own offloading objectives, without sharing that information to any other users or the operator. This also conforms with privacy requirements that might be imposed by the vehicles. In this study, we define six different short and long term, individual and system objectives, and periodically sample objective weights for each user to simulate their changing choice and preference of objectives. With our MARL algorithm, the users learn an optimal offloading strategy and compete for edge-cloud computing resources, despite this frequent change. Our contributions are:

\begin{itemize}
\item To the best of our knowledge, we are the first ones to address the multi-objective nature of ITS applications in its distributed, non-stationary and adversarial environment. Our multi-agent, multi-objective algorithm can optimize frequently changing combinations of objectives and preferences.
\item We train one optimal initial model offline, then deploy the model to each independent agent representing a vehicle user, who is able to change its private objectives and update its offloading strategy through online few-shot learning, needing low retraining cost and no prior knowledge for reward shaping. Our solution outperforms the benchmarking state-of-the-art algorithms on all individual and system metrics. Also, in a heterogeneous environment with different competing algorithms, our algorithm increases bottom-line resource efficiency, such that other algorithms in the environment also benefit from improved offloading rate and fairness.
\item Our algorithm can be modularized and trained asynchronously. We test the runtime inference performance of our algorithm on a single-board computer with a GPU and show that inference in $6$ milliseconds is feasible.
\item We provide public access to our code and data at \cite{moodysource}.
\end{itemize}

Sec.~\ref{sec:related} reviews existing studies on RL for ITS applications such as offloading and resource allocation and multi-objective RL approaches; Sec.~\ref{sec:modelproblem} describes our generic MAS modeled as an auction and formulates the multi-objective optimization problem accordingly; Sec.~\ref{sec:solution} introduces our algorithm; Sec.~\ref{sec:evaluation} analyzes simulation results and Sec.~\ref{sec:chiptest} discusses practical concerns.

\section{Preliminaries \& related work}
\label{sec:related}

RL algorithms are increasingly used to optimize performance in ITS applications such as offloading \cite{ning2020intelligent}\cite{ma2022drl} and resource allocation \cite{zhu2021deep}\cite{liu2023asynchronous}. Due to the distributed nature of the ITS environment, game-theoretic approaches~\cite{bajracharya2021dynamic}\cite{xia2022distributed} and multi-agent systems~\cite{wei2022privacy}\cite{ju2023joint} combined with RL are also common approaches in the recent years. Some studies such as \cite{xu2022computation}, \cite{yao2022dynamic} and \cite{ju2023joint} consider the multi-objective nature of ITS applications, they either decompose the objectives into subproblems \cite{xu2022computation} or convert the problem into a single-objective one through scalarization~\cite{yao2022dynamic}\cite{ju2023joint}. 

All of these studies simplify the ITS environment in some regards: \cite{ma2022drl} and \cite{liu2023asynchronous} consider only one single objective; \cite{zhu2021deep}, \cite{yao2022dynamic} and \cite{liu2023asynchronous} require complete information to centrally solve the optimization problem, but in a dynamic environment, acquiring enough information for a centralized approach is often not feasible or violating user's privacy; \cite{ning2020intelligent} and \cite{xu2022computation} consider only system objectives, not user objectives, and it is assumed that users do not make offloading and resource allocation decisions -- this assumption may not apply to the highly individual and customized environment of ITS, where even today, vehicle and mobile users are participating in offloading and resource allocation decisions in the network, motivated only by their individual objectives. \cite{bajracharya2021dynamic}, \cite{xia2022distributed}, \cite{wei2022privacy} and \cite{ju2023joint} assume all users are cooperative with common objectives, and if users have multiple objectives, the modeling complexity will increase significantly. 

Many real-world decision-making problems consider multiple, sometimes contradicting objectives \cite{cho2017survey}. This is unlike a single-objective problem (SOP) where the objective is scalar and totally ordered: in a multi-objective problem (MOP), the objectives are only partially ordered \cite{shapley1959equilibrium}. An MOP is formulated as finding decision variables that lead to solutions on the Pareto frontier: $f(\mathbf{x})=(f_1(\mathbf{x}), \cdots, f_l(\mathbf{x}))$ s.t. $\mathbf{x} \in \mathbf{K} \subseteq \mathbb{R}^n$, where $f$ is a vector of $l$ objective functions, $\mathbf{x}$ is the decision variable, and $\mathbf{K}$ is the feasible region in an $n$-dimensional decision variable space. Since $f$ can only be partially ordered, we use the \emph{Pareto frontier} to represent a set of equivalent solutions: for a solution on that frontier, no objective can be improved without at least one other objective being worsened. 

Some approaches to solving MOPs extend equilibria concepts to multi-objective settings \cite{patrone2007multicriteria,mouaddib2007towards,perny2013approximation,jonker2017automated}. They all assume some degree of cooperation and communication between agents; they therefore differ from our competitive environment setting where agents do not share information (Sec.~\ref{sec:model}). Reference~\cite{ramos2019budged} assumes a stationary environment, which is different from our multi-state MDP and dynamic environment setting. Reference~\cite{bousia2016multiobjective} assumes complete information that is different from our partial information assumption. Other approaches try to find discrete solutions on a Pareto frontier in stationary environments through objective selection \cite{9055068} or decomposition \cite{8955922, 7390047,9234039}. However, in a dynamic environment like ours, they do not meet the challenge of huge state and action space, unknown state distributions, and MDP with continuing tasks. 

Reinforcement learning (RL) is commonly used to explore huge state and action space, but most RL algorithms only solve SOP \cite{sutton2018reinforcement}. The goal of a single-objective RL algorithm is to maximize the return $J=\mathbb{E}[\sum_{t=0}^\mathbb{T} r_t]$, where $\mathbb{T}$ is the time horizon and $r_t$ is a scalar-valued reward at time $t$, expectation is taken over random rewards. In a multi-objective RL algorithm, we have a vector-valued return for $|O|$ objectives: $\mathbf{J}=\{J_{(o)}|o = \{1,\dots,O\}\}\in \mathbb{R}^{|O|}$, the return for each objective $o$ is $J_{(o)}=\mathbb{E}[\sum_t r_{o,t}]$. Owing to partial order of rewards, such a situation is not directly amenable to standard RL techniques; either it needs to be simplified into an SOP of finding only one of many equivalent optimal solutions (using a constant weight vector $\mathbf{W} \in \mathbb{R}^{|O|}$ to form a single reward $J=\mathbf{W}^T\mathbf{J}$), or the Pareto frontier of all optimal solutions needs to be characterized explicitly.

Even if an MOP can be simplified into an SOP, reference~\cite{hayes2022practical} points out that such simplification requires much theoretical knowledge for reward engineering and manual tuning when objective preferences change over time; a scalarized reward is also inexplainable, i.e., the scalarization does not always reflect the real relationship between decision variables and objectives. Such methods are therefore sensitive to preference changes.  

Instead, multi-objective RL algorithms aim at explicitly finding the Pareto frontier. Such algorithms can be categorized into multiple-model and single-model methods. Multiple-model methods result in multiple, independent models, each aiming to optimize one point on the Pareto frontier, making learning and inference inefficient. They often have high computational cost in high-dimensional objective spaces and are inflexible in a dynamic environment~\cite{parisi2014policy}. Single-model methods such as in \cite{pirotta2015multi} train only one model for all solutions to the MOP, but it is computationally expensive. MAML~\cite{finn2017model} combines the two methods: multiple models are trained for their specific objectives and combined into a generic model, which can be quickly retrained for any new objective with only a few sample data points (\emph{``shots''}). References \cite{kayaalp2022dif} and \cite{AlShedivatBBSM18} extend the method in \cite{finn2017model} from single-agent to multi-agent, but the former considers a stationary environment, and the latter formulates a non-stationary SOP as a stationary MOP. To the best of our knowledge, there have not been studies of multi-agent non-stationary environments with multiple objectives. 
Although \cite{finn2017model} provides a framework for two-phase multi-task training, it does not suggest a choice of the model, as it does not consider any specific problem or application; due to the complexity of the approach, real-life implementation of their method in a dynamic environment such as ITS is problematic. In our study, we design our own multi-agent algorithm for a distributed, dynamic environment with sparse and delayed rewards that is capable of automatically triggering adaptive online retraining (Sec.~\ref{sec:solution}). We also apply various performance improvement measures (Sec.~\ref{sec:chiptest}) to address practical concerns, making our model more suitable for real-life implementation. 

\begin{figure}[t]
	\centering
	{\includegraphics[width=0.8\linewidth]{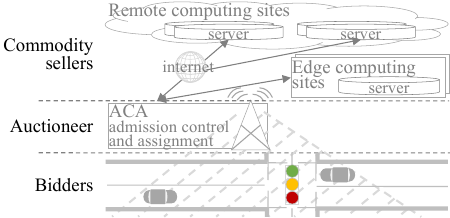}}
	\vspace{-0.2cm}
	\caption{Vehicles request for services from the ACA in range through bidding. The ACA is connected to computing sites both on the edge and in the cloud. The ACA serves as auctioneer, and the computing sites are sellers of resources.}
	\label{topo}
\end{figure}

\section{System model and problem formulation}
\label{sec:modelproblem}

In Sec.~\ref{sec:model}, we introduce our system model for ITS computation resource allocation in edge cloud and show how it can be described as a repeated auction.
In Sec.~\ref{sec:problem}, we formally define the decentralized optimization problem in an auction mechanism. In Sec.~\ref{sec:utility}, we describe the multi-agent system that simulates the mechanism and the interaction.

\subsection{System model}
\label{sec:model}

Our system is designed as an abstraction of the classic edge-cloud computing architecture. An example topology is in Fig.~\ref{topo}. Vehicles get service requests such as image segmentation and motion planning in self-driving applications, each request with its own quality-of-service (QoS) requirements such as deadline. When vehicles cannot process all of the requests on their own onboard units, they try to offload these services to edge-cloud computing sites through road-side units equipped with multi-access edge computing devices. The road-side units are responsible for admission control and assignment (ACA) of service requests.

\begin{figure}[t]
	\centering
	{\includegraphics[width=0.98\linewidth]{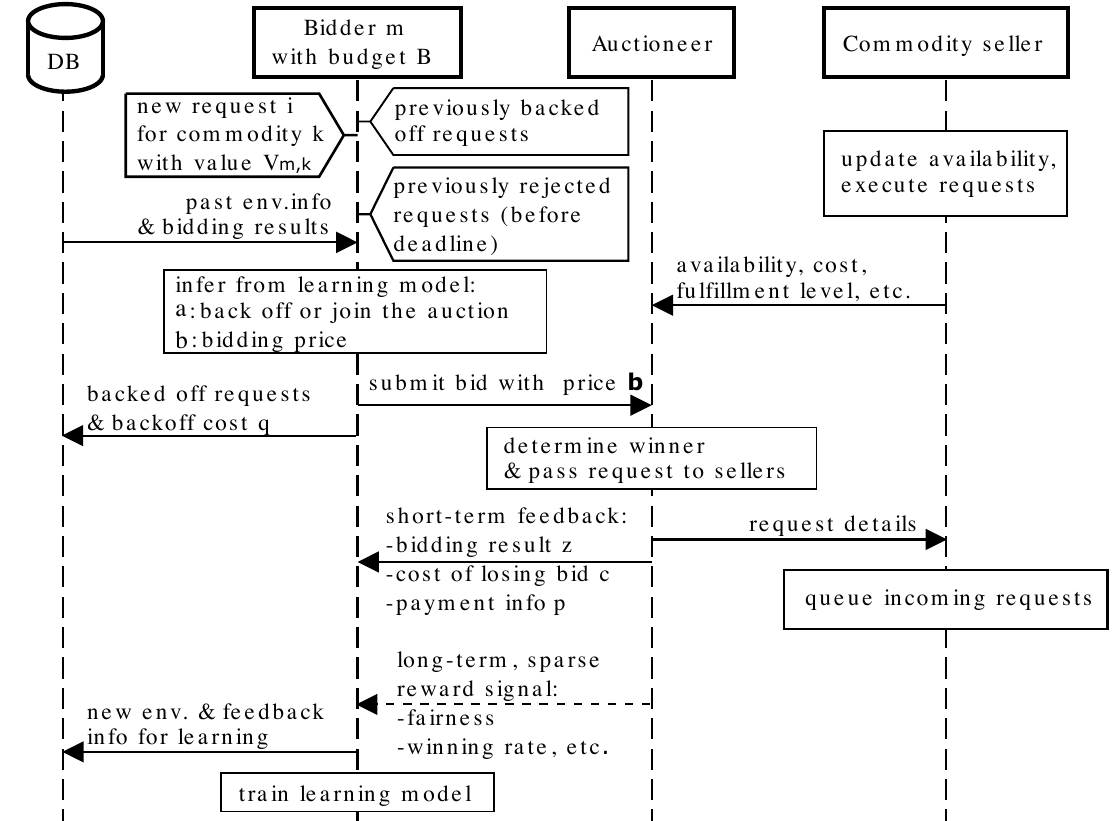}}
	\vspace{-0.2cm}
	\caption{Bidders join repeated auctions with one auctioneer and multiple commodity sellers. A bidder $m$ decides to join with bid $i$ or back off with cost $q$, based on past info and current observations. The auctioneer determines the winner, sends back bidding results $z$ and $c$, payment $p$ and rewards. Commodity sellers execute requests passed on by the auctioneer. Only the bidders can learn.}
	\label{flow}
\end{figure}

Among classic decentralized decision-making mechanisms, we use an auction mechanism because it is most suitable in a dynamic and competitive environment, where the number of bidders and their preferences vary over time, or the bidders' private valuations of the same commodity are very different~\cite{tan2022multi}. Such conditions are typical of V2X environments. 
Specifically, we set up the system as an auction with multiple vehicles as bidders, one ACA as the auctioneer, and multiple computing sites as commodity sellers. A diagram of the mechanism is in Fig.~\ref{flow}. The bidders do not share information with other bidders or commodity sellers; they only communicate with the auctioneer. Our study focuses on the behavior of the independent bidders, conceived of as \emph{agents}. Each bidder has multiple objectives to achieve in the auction. As in real life, the bidder's preference of objectives changes over time, and it needs to learn the Pareto frontier of the MOP to respond quickly to changes \cite{raquel2013dynamic}. We introduce each component of the system below. Table~\ref{tab:problem} summarizes the notation.

\subsubsection{Commodities}

Commodities are typically products or services. In V2X, the commodities are the service slots on the edge cloud for computation offloading. Let $K$ be the set of commodity types. To be executed, each type $k \in K$ has its own specification and resource needs (in terms of material and time). Let $M$ be the set of bidders. Over time, a bidder $m \in M$ gets requests for one or multiple types of commodities and tries to offload and fulfill the request within the specified deadline. All instances of commodities of the same type are equivalent. At time $t$, the bidding price for type $k$ is denoted $b_{m,k}^t$. Each type $k$ has a total of $n_k^t$ available service slots in computing sites. Maximum commodity availability is fixed per type and is not interchangeable between types.

\subsubsection{Bidders} 

A bidder (in V2X, a vehicle) $m$ has private, time-invariant \emph{valuation} $v_{m,k}$ (i.e.\, the benefit it derives from winning the commodity, or in V2X, from successfully offloading its task) for each type $k$ and an initial wealth of $B_m^0$. At time $t$, $m$ can submit its bid denoted $i_k^t \in I$ to the auctioneer, it contains the service request and the bidding price $b_{m,k}^t$. Its direct \emph{payoff} from the auction is $v_{m,k}$ minus its payment to the auctioneer $p^t_{k}$, if it wins the bid, 0 if it loses. Note that the payment to the auctioneer maybe be the same or lower than the bidding price ($p^t_{k} \leq b_{m, k}^t$), depending on the auction type. The bidder's first \textbf{objective $o_1$} is to maximize average \emph{utility}: the payoff minus additional costs due to losing a bid or having to rebid later. In Sec.~\ref{sec:utility}, we further break down $o_1$ into three sub-objectives.

Instead of bidding at time $t$, $m$ also has the option to back off (i.e.\ delay its bid), hoping for less competition for the commodity in the future, but, on the other hand, using up time towards the fixed deadline and thus making the bid more urgent. If the bid passes the deadline, it is viewed as lost (i.e.\, 0 payoff with cost of losing the bid). Specifically, \begin{inparaenum}[1)] \item bidders are incentivized to balance between backoff and immediate bidding; \item backoff time and bidding price can be learned rather than randomly chosen; \item learning is only based on information visible to the bidder.\end{inparaenum}

Besides maximizing utility, the bidder's second individual \textbf{objective $o_2$} is to minimize its long-term offloading failure rate. The bidder can also be incentivized to consider system objectives, as would be detailized in the following subsections. Each bidder's preference over all objectives is private and expressed through a non-negative preference vector $\mathbf{W}_m=\{W_m^{o}| \forall o \in O\}$ that can change frequently over time. The bidder independently learns a bidding strategy to maximize its reward from objective achievements weighted by its preference. We study different learning algorithms in each bidder.  

\begin{table}[t]
 \fontsize{8}{10}\selectfont
 \centering
 \captionof{table}{Problem formulation}
 \label{tab:problem}
 \begin{tabular}{c l c l }
 Sym & Description & Sym & Description\\
 \toprule
 $k \in K$ & commodity type & $n_k$ & $k$'s availability\\
 $i \in I$ & bid/request & $m \in M$ & bidder\\
 $B$ & budget & $v$ & valuation\\
 $\alpha$ & backoff decision & $b$ & bidding price\\
 $c$ & cost of losing the bid & $q$ & backoff cost\\
 $p$ & payment & $z$ & bidding outcome\\
 $u$ & auction utility & $\beta$ & resource utilization\\
 $r$ & reward & $\mathbf{W}$ & preference vector\\
 $o \in O$ & objective & $\pi$ & bidding strategy\\
 \bottomrule 
 \end{tabular}
\end{table} 

\subsubsection{Auctioneer and commodity sellers} 

In our system, we have one auctioneer that determines the winners of auctions. In V2X, the auctioneer is the road-side unit with multi-access edge computing device (MEC), also called the ACA, because it controls admission of service requests from vehicles and assigns admitted requests to edge-cloud computing sites (i.e.\, commodity sellers). It determines the winner through an auction mechanism: service requests with the highest bidding prices are prioritized. The auction is repeated in each discrete time step $t$, if there are active bidders and commodity is available ($n_k^t>0$). The auctioneer passes winning bidders' requests to the commodity sellers with the lowest selling price (e.g.\, set by the commodity sellers according to some load-balancing heuristics). Each bid can only be assigned to one seller. If no seller can fulfill the request, the bids are rejected.

For a rejected bid, the bidder can rebid by a number of times. The auctioneer sets the maximum permitted rebidding times to balance between low bidding failure rate and additional bidding overhead to the system (e.g.\, communication overhead in V2X). For simplicity, we allow rebidding once, the same as in \cite{tan2022multi}. If the bid is accepted by the auctioneer, but not executed by the commodity seller within its deadline, the seller drops the bid and informs the auctioneer and the bidder. Both rejected and dropped bids are considered failures. 

After the auction round, the auctioneer sends the bidders the bidding outcome, payment, and reward signals for system objective achievements, e.g.\, resource utilization, fairness, etc. The bidder also calculates rewards related to its individual objectives, such as bidding failure rate. It can decide whether to use auctioneer's reward signals for learning. For example, one bidder may have low preference for system objectives and ignore the information; another one may find the information useful. The auctioneer has the \textbf{objective $o_3$} to maximize overall fairness among bidders.

The commodity sellers dynamically adjust their commodity prices. Since our study focuses on the behavior of the bidders, we make the sellers passive (i.e., not learning-capable) and use a simplified pricing heuristic with load-balancing effect: a seller with higher percentage of unsold resources sells at a lower price; the requests are assigned to the seller with the lowest price, until all sellers have similar utilization and price. The commodity sellers have the \textbf{objective $o_4$} to maximize system utilization and minimize variance in utilization. With low variance, commodity sellers can better plan long-term resource availability, reaching high utilization especially in high contention, saving cost while keeping the same service level.

We call $o_1$ and $o_2$ a bidder's individual objectives, $o_3$ and $o_4$ the system objectives. Although the auctioneer and commodity sellers cannot force the bidders to consider system objectives, in Sec.~\ref{sec:eval} we show that the reward signals help bidders learn the correlation between individual and system objectives, and by considering system objectives, the bidders effectively earn higher reward on their individual objectives.

To make our system resemble the fast-paced edge-cloud computing architecture in real life, we add transmission delay between bidders, the auctioneer and commodity sellers, and randomize resource requirement, queuing time and processing time for request execution. Each bidder learns its optimal bidding strategy despite noisy state information in such a dynamic environment. Sec.~\ref{sec:evaluation} describes the simulated V2X environment and its example self-driving applications in detail.

\subsection{Problem formulation}
\label{sec:problem}

We now formulate the distributed decision making problem related to the system model in Sec.~\ref{sec:model} as an auction for multiple commodities. From its bidding strategy $\pi_m$, bidder $m$ draws its actions $\mathbf{\alpha}_m^t =\{\alpha^t_{m,k} \in \{0,1\}\}$ and $\mathbf{b}_m^t =\{b^t_{m,k} \in \mathbb R_+\}$ for each service type. $\alpha$ is the vector of backoff decisions, $\mathbf{b}$ is the vector of bidding prices. More specifically, bidder $m$'s options for each bid are: \begin{inparaenum}[1)] \item back off ($\alpha^t_{m,k}=0$) with a backoff cost $q^t_{m,k}$, or \item bid ($\alpha^t_{m,k}=1$) with  price $b^t_{m,k}$. \end{inparaenum} To avoid overbidding, at any time $t$, $\sum_{k} \alpha^t_{m,k} b^t_{m,k} \leq B_m^t$.

From bidder $m$'s perspective, the competing bidders (denoted $-m$) draw their actions from a joint strategy distribution $\pi_{-m}^t$ that is an unknown function of $(\mathbf{p}^1,\cdots, \mathbf{p}^{t-1})$, where $\mathbf{p}^t \in \mathbb R_+^{|K|}$ is the vector of final prices at the end of time $t$. All bidders get the vector of commodity prices for time $t$, denoted $\mathbf{p}^t$, as feedback from the auctioneer. If bidder $m$ wins its bid $i_k^t$ indicated by bidding outcome $z_{m,k}^t=1$, it pays $p_{k}^t$ to the auctioneer. If the bidder loses (i.e.\, $z_{m,k}^t=0$), it pays 0 to the auctioneer, but it also has a cost associated with losing the bid, denoted by $c_{m,k}^t$. If rebidding is permitted and $i_k^t$ has not reached its deadline, $m$ repeats the decision-making process in $t+1$. If $i_k^t$ passes the deadline before it is admitted, it is viewed as a lost bid with cost $c_{m,k}^t$.

The auction repeats for $\mathbb{T}$ rounds, in every auction round, bidder $m$'s utility is $u_m^t(\alpha_m^t,\mathbf{b}_m^t,\mathbf{p}^t,\mathbf{z}_{m}^t,\mathbf{c}_m^t,\mathbf{q}_m^t)$, the utility is added to the wealth pool: $B_m^{t+1} = B_m^t + u_m^t$. If $B_m^{t+1}\leq 0$, bidder $m$ loses all the unfinished bids with cost $c_{m,k}^t$ and is reset. 

Next, we formulate the problem with multiple objectives $o \in O$ as described in Sec.~\ref{sec:model}. Bidder~$m$ receives a reward vector $\mathbf{r}_m^t \in \mathbb{R}^{|O|}$ in random intervals from its own observation and the feedback signals from the auctioneer for its achievement of these objectives. More details will be provided in the next section. Each bidder's preference vector over the objectives is $\mathbf{W}_m^t \in \mathbb{R}^{|O|}$. Bidder preferences can change over time. In real life, changes in preference can be driven by long-term shifts in societal, legal and personal attitudes, or short-term private prioritization, etc. The bidder's goal is to maximize expected return $\mathcal{J}_m =\frac{1}{\mathbb{T}} \sum_{t=1}^\mathbb{T} (\mathbf{W}_m^t)^T \cdot \mathbf{r}_m^t,\mathbb{T} \to \infty$, where $\mathbf{W}_m^t$ is time-variant and unknown to the bidder in advance.

Typical RL techniques learn to maximize reward with a constant preference vector over the multiple objectives. This is essentially one single point on the Pareto frontier of the MOP. Our approach in Sec.~\ref{sec:solution} finds the shape of the Pareto frontier, befitting the V2X environment where vehicles have time-variant preference vectors $\mathbf{W}_m^t$ that is unknown in advance.

In the following Sec.~\ref{sec:utility}, we build a MAS to simulate the auction mechanism and bidders with multiple objectives. 

\subsection{Multi-agent system (MAS) for the auction mechanism}
\label{sec:utility}

We design a MAS where each auctioneer and commodity seller is represented by a passive agent. Each bidder $m$ is one active (i.e., learning-capable) agent. Other bidders are denoted $-m$. Thus, the agents in our MAS represent actual entities in V2X (Fig.~\ref{topo}).

For simplicity, we omit the notation for time step $t$ in this section. For each commodity type $k$, the bidder is given a private valuation $v_{m,k}$ that is \begin{inparaenum}[1)] \item linear to the bidder's estimated resource needs for the commodity and \item within its initial wealth $B_m^0$. \end{inparaenum} \cite{tan2022multi} proves the first condition guarantees Pareto optimality in the case the bidder has a constant preference vector over all objectives, and the second condition avoids overbidding under rationality. We do not consider irrational or malicious bidders, e.g., whose goal is to reduce social welfare even if their individual outcome may be hurt.  Bidder $m$ decides whether it backs off for the current auction round ($\alpha_{m,k}=0$) or bids with price $b_{m,k}$. The auctioneer receives only the bidders' required commodity type $k$ and bidding price; at the end of each auction round, the auctioneer sends back to $m$ the bidding outcome $z_{m,k}$, the final price (i.e., required payment from winning bidders)  $p_k$, as well as system rewards such as fairness score and resource utilization $\beta$ if they are available.

Among different types of auctions, we choose to use the second-price auction. Because second-price auctions maximize social welfare (i.e., total utility of all bidders) instead of auctioneer profit, it is commonly used in auctions for public goods. For $n_k=1$, the required payment $p_k$ is the price of the second highest bid (hence the name ``second-price auction''); for  $n_k>1$ available commodities, this would be the $n_k^{\textrm{th}}$ highest bid, denoted $b^{*}_k$. For the winning bidders with the highest $n_k$ bids (ties are randomly broken), $z_{m,k}=1$ and $p_{k} = b_k^*$. 

In each auction round, bidder $m$ has \textbf{objective~$o_1$}: maximize immediate auction utility $r_m^{o_1}=u_m$. Objective~$o_1$ is broken down into three sub-objectives. \begin{inparaenum}[1)] \item $o_{1-1}$: maximize payoff $r_m^{(k,o_{1-1})}=\alpha_{m,k} \cdot z_{m,k} \cdot (v_{m,k}-b_k^*)$. \item $o_{1-2}$: minimize the chance of being rejected by the auctioneer. The cost of bidding and then losing the bid is $r_m^{(k,o_{1-2})}=-\alpha_{m,k} \cdot (1-z_{m,k}) \cdot c_{m,k}$. \item $o_{1-3}$: minimize backoff time. If backed off, $m$ has cost $r_m^{(k,o_3)}=-(1-\alpha_{m,k}) \cdot q_{m,k}$. \end{inparaenum}

\begin{flalign}\label{eq:rewardbackoff}
\nonumber u_{m,k} =& r_m^{(k,o_{1-1})} + W_{m}^{o_{1-2}} r_m^{(k,o_{1-2})} + W_{m}^{o_{1-3}} r_m^{(k,o_{1-3})}\\
r_m^{o_1}=& u_m = \sum_{k \in I} u_{m,k}
\end{flalign}

The cost terms $c_{m,k}$ and $q_{m,k}$ with preferences $W_m^{o_{1-2}}$ and $W_m^{o_{1-3}}$ quantify tradeoff between long backoff time and risky bidding. In our implementation (Sec.~\ref{sec:evaluation}), $c_{m,k}=v_{m,k}$, $q_{m,k}$ is reciprocal to the time-to-deadline, and non-negative weights $W_m^{o_{1-2}}+W_m^{o_{1-3}}=1$. Sec.~\ref{sec:sensitivity} shows our algorithm is not sensitive to changes in the hyperparameters $v_{m,k}$ and $q_{m,k}$.

Bidder $m$'s long-term individual \textbf{objective $o_2$} is to minimize $r_m^{o_2}=\textrm{OFR}_m \in (0,1)$ at long intervals with preference $W_m^{o_2}$. Long-term objectives are only available to $m$ at the end of each interval. The short-term system \textbf{objective $o_3$} of maximizing resource utilization $r_{m}^{o_3}=\beta$ and the long-term system \textbf{objective $o_4$} of maximizing fairness $r_m^{o_4}=\textrm{Fairness}$ are the same for all bidders and broadcasted to all. Bidders do not have to respect the system objectives; their preferences are reflected in the values $W_m^{o_3}$ and $W_m^{o_4}$. A preference of $0$ means the bidder does not consider system objectives at all. The definitions of $\textrm{OFR}$, $\beta$ and $\textrm{Fairness}$ for our implementation are in Sec.~\ref{subsec:setup}.

The reward for objective achievement in each time step is: 

\begin{flalign}\label{eq:extrinsicrewarddef}
r_{e,m} = W_m^{o_1} \cdot r_m^{o_1} + W_m^{o_2} \cdot r_m^{o_2} +W_m^{o_3} \cdot r_m^{o_3} + W_m^{o_4} \cdot r_m^{o_4}
\end{flalign}

\noindent The notation $r_{e,m}$ is for the scalarized \emph{extrinsic reward} (Sec.~\ref{sec:adaptivecreditassignment}) specific for the bidder $m$ with preference vector $\mathbf{W}_m^t$ at time $t$. We let $W_m^{o_1}+W_m^{o_3}=1$ to balance the short-term objectives, and $W_m^{o_2}+W_m^{o_4}=1$ to balance the long-term objectives. 

Next, we propose an algorithm that learns to maximize $r_{e,m}$ over time, with changing $\mathbf{W}_m^t$.

\section{Proposed solution}
\label{sec:solution}

To solve the long-term, multi-objective reward maximization problem, we propose MOODY: \textbf{M}ulti-\textbf{O}bjective \textbf{O}ptimization through \textbf{D}istributed reinforcement learning with dela\textbf{Y}ed reward. 

The algorithm's multi-objective design allows users to weigh their own objectives without information sharing with the operator. 

\begin{figure}[t]
	\centering
	\includegraphics[width=0.95\linewidth]{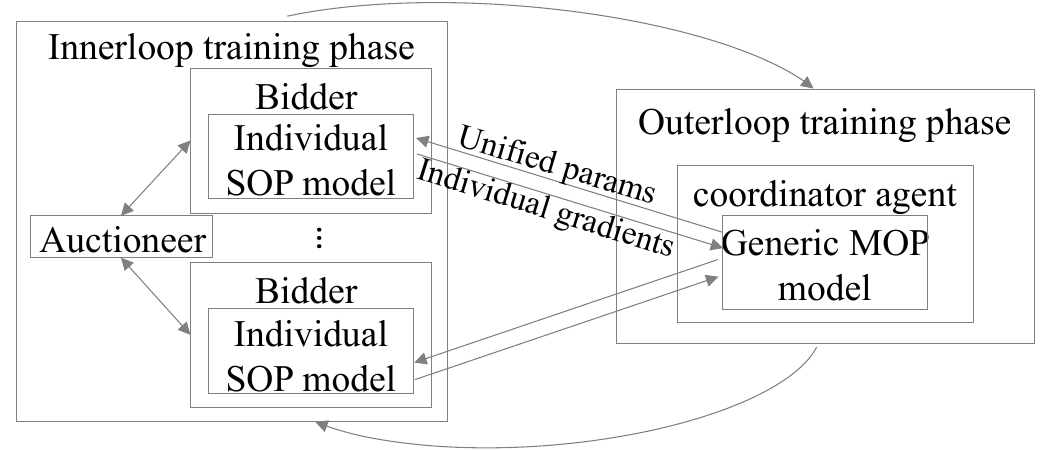}
	\vspace{-0.2cm}
	\caption{Two-phase offline training}
	\label{metatraining}
\end{figure}

In this study, we define six different short and long term, individual and system objectives, and periodically sample objective weights for each user to simulate their changing choice and preference of objectives. If the sampled preference weight is 0, the user does not choose that objective. 

Our technique comprises two parts: the offline training cycle and the online inference/retrain cycles. In the offline training cycle, the approach is further split into two phases: the inner-loop training phase and the outer-loop training phase. Authors of \cite{finn2017model} demonstrated that such a two-phase training method creates a initial generic MOP model that can be easily retrained online for a different task.

In the inner-loop training phase, a local agent trains with a uniform randomly sampled, constant preference vector and tries to find an optimal solution on its Pareto frontier for the given preference vector.

In the outer-loop training phase, one coordinator agent combines all results from inner-loop trainings. The inner-loop and outer-loop training happens alternatively (Fig.~\ref{metatraining}). At the end of the training cycle, we have an initial generic model that, given any new preference vector, can infer an action which leads to a set of rewards that is close to the Pareto frontier.

In our case, the individual long-term objective of OFR and the system long-term objective fairness are positively correlated to each other (see Sec. \ref{sec:eval}), therefore all users' optimal solutions regardless of their preference vectors would have similar (maximum) rewards in the two objectives. In Sec.~\ref{sec:sensitivity}, we show with results from the inference/retrain cycle how the rewards for these objectives are not sensitive to vehicle users' different and changing preferences. Hence, although each user makes independent decisions based on private objectives and preferences, their collective decision-making results in a better solution for all.

Based on the two-phase training framework of \cite{finn2017model}, we make several improvements: \begin{inparaenum}[1)] \item we design a specific inner-loop algorithm for our multi-agent application scenario, it outperforms classic RL algorithms such as actor-critic in dynamic environments with sparse and delayed rewards. \item In the outer loop, we implement the parallel stochastic gradient ascent method~\cite{lian2018asynchronous} using fully distributed, asynchronous federated learning to increase learning efficiency. \item We propose an adaptive online retraining method that continuously predicts long-term reward; a decreasing prediction accuracy triggers a short, few-shot online retraining cycle. Our model is therefore more adaptive to changing environment and objectives compared to \cite{finn2017model}, which only retrains the model at the beginning of deployment. \end{inparaenum}

The two-phase training cycle takes place offline with gradient-sharing between the generic model and the local, single models. Otherwise, observation data, hyperparameters for initialization and objective preferences remain private to the local agents. Once the training cycle is over, we reset the simulation environment to have all local agents initialized with the extensively trained generic model, then test them for online inference and retraining without further parameter sharing, in a realistic test environment. 

Section~\ref{sec:adaptivecreditassignment} introduces the inner-loop RL algorithm. Section~\ref{sec:federated} describes the parallel stochastic gradient ascent method using asynchronous federated learning in the outer-loop offline training phase. Section~\ref{sec:adaptivecreditassignment2} introduces our adaptive retraining method in the online retraining cycle. Notations are in Table~\ref{tab:algorithm}.

	\begin{table}[t]
     \fontsize{8}{10}\selectfont
	 \centering
	 \captionof{table}{Proposed solution}
	 \label{tab:algorithm}
	 \begin{tabular}{c  l  c  l}
	 Sym & Description & Sym & Description\\
	 \toprule
	 $\theta$ & model parameters & $\Gamma$ & learning rate\\
	 $\gamma$ & discount rate & $\delta$ & TD error\\
	 $A$ & action & $S$ & state\\
	 $V$ & state value & $\pi$ & target policy\\
	 $\mathcal{J}$ & scalarized return & $\tau$ & inner-loop training shots\\
     $r_{e}$ & extrinsic reward & $r_{i}$ & intrinsic reward\\  
     $\zeta$ & best response strategy & $\psi$ & behavior strategy\\
     $\eta$ & best response weight & $L_f$ & state prediction loss \\
     $\phi$ & state features & $\epsilon$ & credit weight for actions\\
	 \bottomrule 
	 \end{tabular}
	\end{table}

\subsection{RL in the inner loop}
\label{sec:adaptivecreditassignment}

In the inner loop, each bidder (local agent) learns autonomously to maximize its reward. The inner-loop algorithm is based on our previous work \cite{tan2022learning}, we change it to suit our multi-objective problem, such that it can now learn to optimize multiple short and long-term objectives with a preference vector $\mathbf{W}_m^t \in (0,1)^{|O|}$ that changes over time (in our simulation, it is drawn from a uniform distribution at random interval). 

State vector $S_m^t$ in the inner loop consists of: \begin{inparaenum}[1)] \item information of $m$'s bids (in our simulated V2X scenario in Sec. \ref{sec:model}, this includes the type of service request, its deadline, resource amount required, etc.); \item limited environment information $m$ gets from the auctioneer, e.g., number of bidders in the network, system utilization, etc.; \item other private bidder conditions such as previous wealth $B_m^{t-1}$; \item previous competitor state $P_{-m}^{t-1}$, represented by previous payments: $P_{-m}^{t-1} = \mathbf{p}^{t-1} = \{p_k^{t-1}| k \in K \}$; \item previous extrinsic reward $r_{e,m}^{t-1}$ as defined in Eq.\ref{eq:extrinsicrewarddef}. \end{inparaenum} We specifically limit the model input to information that is easily obtainable by the bidder. This meets real-life requirements for limited information-sharing between bidders. Through feature extraction layers, we get feature vector $\phi_m^t$ from stacked state vectors from the memory as input. Output from each model is the bidding strategy for time step $t$, including backoff decision $\alpha_m^t$ and bidding price $\mathbf{b}_m^t$. 

\begin{figure}[t]
	\centering
	\includegraphics[width=0.95\linewidth]{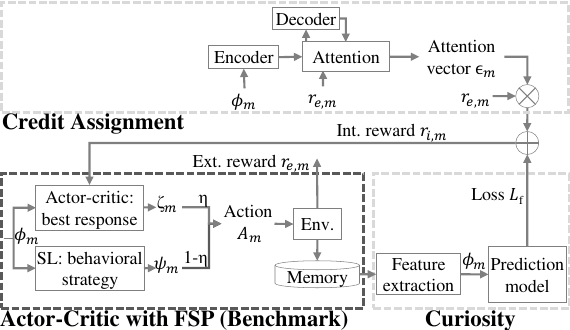}
	\vspace{-0.2cm}
	\caption{Inner loop RL}
	\label{innerlooprl}
\end{figure}

The local, single models have the same structure as the generic model. It consists of three parts (Fig.~\ref{innerlooprl}): \begin{inparaenum}[1)] \item a fictitious self-play (FSP) module \cite{heinrich2015fictitious}, including an RL with actor-critic and a supervised learning (SL) part, \item a curiosity-learning \cite{pathak2017curiosity} module, and \item a credit-assignment module. \end{inparaenum} 

With the FSP method, an RL learns a bidding strategy that is the best response to other bidders' actions; parallel to the RL, the SL learns a behavioral strategy from the bidder's own past bidding behaviors, disregarding current competitor state, and the final bidding decision is selected between the best-response and the behavioral strategy with a factor $\eta \in (0,1)$ that increases over time. This stabilizes learning in a dynamic environment and improves the overall convergence property \cite{heinrich2015fictitious}. In Sec.~\ref{sec:eval}, we use a stand-alone actor-critic (AC) with FSP as one of the benchmark algorithms. The curiosity module \cite{pathak2017curiosity} extracts state features that directly influence agent action and disregards less useful state information, thus improving the model's ability to generalize. It predicts next state and inserts the prediction loss as intrinsic reward $r_{i,m}^t$ between sparse extrinsic rewards $r_{e,m}^t$ to encourage exploration in unfamiliar state space. Finally, the credit assignment module predicts and breaks down long-term, delayed rewards and attributes them to short-term actions through a weight vector from the attention layer. It uses an RNN and updates parameters at long intervals (i.e., when long-term reward is available). 

In an ablation study in \cite{tan2022learning}, we compared the contribution of each module to the agent's performance. We simulated two common types of repeated auctions with a single commodity and pitched three algorithms against each other: \begin{inparaenum}[1)] \item a stand-alone FSP with the basic AC algorithm, \item the AC plus curiosity learning, and \item the AC plus both curiosity learning and credit assignment. \end{inparaenum} Results showed that with each additional module, the performance became better, and the combination of three modules outperformed all others.
Based on this result, we use all three modules in this study.

\begin{algorithm}[t]
  	\small
	\begin{algorithmic}[1]
	\STATE Initialize $T=0$, $B_m^0$, $\mathbf{v}_m$, $\Gamma_m$, $\gamma_m$, $\eta_m$, $\tau$.
	\WHILE{true}
		\STATE $t \gets T+1$, receive new preferences $\mathbf{W}_m^t$ if available.
		\STATE Receive $\theta^0$ from coordinator agent, initialize $\theta_m=\theta^0$.
		\WHILE{$t \leq T+\tau$}
                \STATE \textbf{Observe and remember:}
                \begin{ALC@g}
                    \STATE Get new service request and add to pipeline at time $t$.
                    \STATE Observe environment variables and past payments.
                    \STATE Retrieve details of all requests in current pipeline.
                    \STATE Create state vector $S_m^t$ and add to memory.
                    \STATE Infer $\phi_{m}^t,L_{fm}^t$ from \emph{curiosity}.
                    \STATE Infer $\epsilon_m^t$ from \emph{credit assignment}.
                    \STATE With $\epsilon_m^t$,  update backwards past $r_{i,m}$'s in memory.
                \end{ALC@g}
                \STATE \textbf{Take action:}
                \begin{ALC@g}
                    \STATE Infer actions $\alpha_{m}^t$,$\mathbf{b}_{m}^t$ from \emph{actor-critic RL with FSP}.
                    \STATE Collect all bids with backoff decision $\alpha=0$:
                    \begin{ALC@g}
                        \STATE Calculate backoff cost $\mathbf{q}_m^t$, update $r_{e,m}^t$ in memory. 
                        \STATE Add those before deadline to pipeline at $t+1$. 
                        \STATE Drop the rest as lost bids with penalty $\mathbf{c}_m^t$.
                    \end{ALC@g}
                    \STATE Submit bids with $\alpha=1$, with prices $\mathbf{b}_{m}^t$.
                \end{ALC@g}
                \STATE \textbf{Collect rewards:}
                \begin{ALC@g}
                    \STATE Observe bidding results $\mathbf{z}_{m}^t$ and payments $\mathbf{p}^t$.
                    \STATE Collect all lost bids with $z=0$:
                    \begin{ALC@g}
                        \STATE Calculate penalty $\mathbf{c}_m^t$, update $r_{e,m}^t$ in memory. 
                        \STATE Collect requests before deadline and can rebid:
                        \STATE Add those to pipeline at $t+1$; drop the rest.
                    \end{ALC@g}
                    \STATE Collect all won bids with $z=1$:
                    \begin{ALC@g}
                        \STATE Calculate auction utility, update $r_{e,m}^t$ in memory.
                        \STATE Update $B_m^t$.
                    \end{ALC@g}
                    \STATE Get other ext. rewards, update $r_{e,m}^t$ in memory.
                    \STATE Calculate and add $r_{i,m}^t$ in memory. 
                \end{ALC@g}
                \STATE \textbf{Update learning model:}
                \begin{ALC@g}
                    \STATE Train \emph{actor-critic RL with FSP} and \emph{curiosity}.
                    \STATE Train \emph{credit assignment} if long-term reward available.
                    \STATE Update $\theta_m$ with gradient $\nabla_{\theta_m^t} \mathcal{J}_m=\delta_m \nabla_{\theta_m} \ln \pi_m$.
                \end{ALC@g}
			  \STATE $t \gets t+1$
		\ENDWHILE
		\STATE Pass $\nabla_{\theta_m^\tau} \mathcal{J}_m$ to coordinator agent.
		\STATE $T \gets T+\tau$.
	\ENDWHILE
  \end{algorithmic}
  \caption{Offline innerloop training of local agent $m$}
  \label{innerloopalg}
  \end{algorithm}

In the beginning of every inner-loop offline training phase, all local agents are initialized with the same generic model that is the outcome of the previous outer-loop phase, with the parameters $\theta^{0}$. During inner-loop training, local agent $m$ trains its own local, single model and does not share parameters or private observations with other agents. At each time step $t$, it receives extrinsic reward $r_{e,m}^t$, including the long-term rewards if they are available. $m$'s curiosity module predicts next state with prediction loss $L_{fm}^t$, and the credit assignment module outputs attention vector $\epsilon_m^t$. The resulting intrinsic reward is $r_{i,m}^t = \epsilon_m^t r_{e,m}^{t} + L_{fm}^t, t \in \{1,\cdots,\tau\}$. The expected return is now $\mathcal{J}_m=\frac{1}{\mathbb{T}} \sum_{t=1}^\mathbb{T} r_{i,m}^t,\mathbb{T} \to \infty$. In trying to maximize $\mathcal{J}_m$, the local agent encourages \begin{inparaenum}[1)] \item actions that bring higher extrinsic reward, \item exploration in less visited states with poor prediction accuracy (high $L_{fm}^t$), and \item actions that contribute more to the accurate prediction of long-term rewards (high $\epsilon_m^t$) \end{inparaenum}. The update rule for $m$'s individual parameters in the inner-loop offline training phase is \cite{sutton2018reinforcement}: 

$\begin{cases}
\theta_m^{t} \gets \theta_m^{t-1} + \Gamma_m \delta_m^{t-1} \nabla_{\theta_m^{t-1}} \ln \pi(A|\phi_m^{t-1},\theta_m^{t-1})\\
\theta_m^{0}=\theta^{0},\forall t\in \{1,\dots,\tau\}\\
\end{cases}$

\noindent where $\delta_m^{t-1}= \mathbf{r}_{i,m}^{t-1}+\gamma_m \hat{V}(\phi_m^t,\theta_m^{t-1})-V(\phi_m^{t-1},\theta_m^{t-1})$ is the TD error, $\Gamma_m$ is the learning rate, and $\gamma_m$ is the discount rate. In our case, action $A=(\alpha_m^t,\mathbf{b}_m^t)$. At the end of $\tau$ shots, the local gradients are passed to the coordinator agent before the next outer-loop phase. The inner-loop algorithm is in Alg.~\ref{innerloopalg}. 

\subsection{Federated learning in the outer loop}
\label{sec:federated}

While independent local agents with the local models learn for $\tau$ shots, the coordinator agent with the generic model waits with the original parameters $\theta^{0}$, until the next update in the outer-loop phase (Fig.\ref{metatraining}). In the outer-loop phase, the goal of the coordinator agent is to maximize all local agents' sum of returns: $\mathcal{J}=\sum_m \mathcal{J}_m(\theta_m^{\tau})$. After $\tau$ shots, at the end of the previous inner-loop phase, the generic model's parameters are $\theta^{0}$, and it uses the local models' gradients to update its parameters: $\theta^{0'} = \theta^{0}+\Gamma\nabla_{\theta^{0}} \mathcal{J}$. Since each individually updated parameter $\theta_m^t, \forall t\in \{1,\dots,\tau\}$ is a function of $\theta^{0}$, using chain rule, the generic model's parameter update is: 

$\theta^{0'}=\theta^{0} + \sum\limits_m \Big{(}\nabla_{\theta_m^{\tau}} \mathcal{J}_m(\theta_m^{\tau}) \prod\limits_{t=1}^{\tau-1} \big{(} \mathcal{I} - \Gamma_m \nabla^2_{\theta_m^{t}} \mathcal{J}_m(\theta_m^{t}) \big{)} \Big{)}$

\noindent where $\mathcal{I} $ is the identity matrix. Although it is computationally expensive, it can be approximated by a first-order derivative with the assumption that both $\Gamma$ and $\tau$ are small \cite{finn2017model,nichol2018first}. $\theta^{0'} = \theta^{0}+\sum_m \Gamma \delta_m^\tau \nabla_{\theta_m^{\tau}} \ln \pi_m^\tau (\theta_m^{\tau})$ is the simplified update rule. Our setup meets the assumptions with $\Gamma= 0.1$ and $\tau= 3$. 

We use asynchronous federated learning to implement the parallel stochastic gradient ascent method (Sec.~\ref{sec:chiptest}). It does not require all local models to be trained and updated at the same time: 
each model is trained based on the availability of new local data. Whenever the local model finishes training for $\tau$ shots, the local agent transmits the gradients to the coordinator agent and gets updated model parameters from it. This reduces data rate needed for gradient and parameter communication and further increases learning efficiency. 

\subsection{Adaptive online retraining}
\label{sec:adaptivecreditassignment2}

After the offline training, and once the model is deployed in a real-world setting, the credit-assignment module continuously predicts rewards. The current reward prediction accuracy is compared to the moving average of past $N$ prediction accuracies, if it falls below the past average, a short $n$-shot retraining cycle is triggered. In our simulation, we use $N=10$ and $n=1$. The algorithm is described in Alg.~\ref{adaptivealg}.

\begin{algorithm}[H]
  	\small
	\begin{algorithmic}[1]
	\STATE Initialize $t=0$, $B_m^0$, $\mathbf{v}_m$, $\Gamma_m$, $\gamma_m$, $\eta_m$, $\tau$, and moving average period $N$ of \emph{credit assignment}'s prediction loss.
     \STATE Initialize with $\theta$ from coordinator agent.
	\WHILE{true}
		\STATE $t \gets t+1$, receive new preferences $\mathbf{W}_m^t$ if available.
           \STATE \textbf{Observe and remember}.
           \STATE \textbf{Take action}.
           \STATE \textbf{Collect rewards}.
           \IF{long-term reward is available}
               \STATE Calculate and store prediction loss of \emph{credit assignment}.
               \IF{current prediction loss > past $N$ average}
                   \STATE \textbf{Update learning model}.
               \ENDIF
            \ENDIF
	\ENDWHILE
  \end{algorithmic}
  \caption{Online adaptive retraining of local agent $m$}
  \label{adaptivealg}
  \end{algorithm}

Simulation results in Sec.~\ref{sec:eval} show the effectiveness of this adaptive online retraining approach. Sec.~\ref{sec:chiptest} mentions practical considerations in online retraining.

\section{Evaluation}
\label{sec:evaluation}

\subsection{Simulation setup}
\label{subsec:setup}

The evaluation has two cycles: offline two-phase training, and online testing / retraining. The coordinator agent with the generic model is only present in the training cycle, it collects gradients from all local agents and learns a generic model. Once deployed in the test environment, all agents are initialized with the same generic model, but then diverge from it by adapting to the environment through online retraining. All agents are independent bidders with private observations and model parameters that are not shared with any other agent. In both cycles, we consider a V2X system as defined in Sec.~\ref{sec:model}: vehicles are bidders who request networking services (commodities); road-side unit or base station acts as auctioneer that controls admission of service requests and assigns them to different computing sites (commodity sellers), which own resources and execute services~\cite{whaiduzzaman2014survey}. Many V2X use cases~\cite{5gaausecase2} can be mapped to this setup.

We develop a Python discrete-event simulator based on the available open-source code \cite{dracosource2}. It is a realistic V2X setup modeled as a 4-way traffic intersection (Fig.~\ref{topo}), with varying number of vehicles of infinite lifespan, one MEC system with ACA and edge computing site, and one remote computing site with non-negligible delay to/from the intersection in data transmission and state information update. The commodity types in the auction correspond to service request types in V2X. We specifically model two self-driving applications: motion planning ($F1$) and image segmentation ($F2$). The details are in Table \ref{tab:setupdiff}. The commodity instances being auctioned are service slots for the different service request types, provided by the computing sites. All environment parameters are randomized to imitate noise in real life. ACA assigns admitted requests to computing sites based on a load-balancing heuristic named resource-intensity-aware load-balancing (RIAL) \cite{8006307}. The method achieves dynamic load-balancing among computing sites through resource pricing that is correlated to the site's load, and loads are shifted to ``cheaper'' sites. The mobility data is generated from SUMO \cite{behrisch2011sumo}, with varying vehicle speed, arrival rate, traffic light phases, etc.

The bidders have the following objectives:

\begin{itemize}
\item Maximize individual short-term (immediate) auction utility: as defined in Sec.~\ref{sec:utility}. 
\item Maximize system short-term resource utilization: load-balancing effect is achieved by encouraging bidding at time of low system utilization \cite{tan2022multi}. Resource utilization is the ratio of resources effectively utilized at computing sites at the time of ACA admission.
\item Minimize long-term individual offloading failure rate (OFR): average ratio of offloading requests that cannot be serviced before deadline. In fact, OFR should include all failed service executions at computing sites until deadline, rather than only those dropped by the bidder or rejected by the ACA. However, this means feedback of bidding result to the bidders is delayed, and the length of delay is specific to each service request. To simplify, we use rejection rate as a proxy to OFR. This is justified by the fact that our system responsiveness (i.e., the ratio of successfully executed requests to all accepted requests) is ca. $99\%$.
\item Maximize long-term system fairness: we use J-index \cite{jain1984quantitative} of payments over the last $\mathcal{T}$ time steps: $\text{Fairness} = \cfrac{(\sum_{m}\sum_{t-\mathcal{T}}^t p_m)^2}{|M|\sum_m (\sum_{t-\mathcal{T}}^{t} p_m)^2},\forall m \in M$. It is commonly used to measure fairness in networking, it is also the reciprocal of the normalized Herfindahl–Hirschman Index \cite{rhoades1993herfindahl}.
\end{itemize}

\begin{table*}[t]
 \fontsize{8}{10}\selectfont
 \centering
 \begin{minipage}{\textwidth}
 \captionof{table}{Setup differences}
 \label{tab:setupdiff}
\centering
 \begin{tabular}{ >{\centering}p{0.2\textwidth}  >{\centering}p{0.32\textwidth}  >{\centering\arraybackslash}p{0.32\textwidth} }
 Environment Parameters & Training Setup & Test Setup\\
 \toprule
service request type & \multicolumn{2}{c}{\makecell{F1: $80$ resource units (abstracted from CPU and memory usage) needed within $100$ time steps (milliseconds)\\F2: $80$ resource units needed within $500$ time steps}}\\
\midrule
service arrival rate & \multicolumn{2}{c}{F1: every $100$ time steps; F2: every $500$ time steps}\\
\midrule
data size & \multicolumn{2}{c}{uplink: F1: $0.4$ Mbit, F2: $4$Mbit; downlink: F1: $0$ (negligible), F2: $0.4$ Mbit}\\
\midrule
latency & \multicolumn{2}{c}{802.11ac: 65m radius, maximum channel width $1.69$ Gbps, throughput=$-26 \times \text{distance} + 1690$ Mbps \cite{shah2015throughput}}\\
\midrule
\makecell{computing site capacity\\(resource units per time step)} & $60$ (low contention) & $10$ (high contention)\\
\midrule
vehicle arrival rate & $1$ every $2.2$ seconds & $1$ every $1$ second\\
\midrule
vehicle speed & $10$ km/h when driving & $30$ km/h\\
\midrule
vehicle count & $22-29$ and slow-changing & $14-30$ and bursty\footnote{We regulate burstiness by adjusting vehicle speed, arrival rate and traffic light phases}\\
 \bottomrule 
 \end{tabular}
\end{minipage}
\end{table*}

	\begin{figure*}[t]
		\centering
		\subcaptionbox{Average RL reward of all bidders increases over time.\label{rl_int_reward}}{\includegraphics[width=0.24\linewidth]{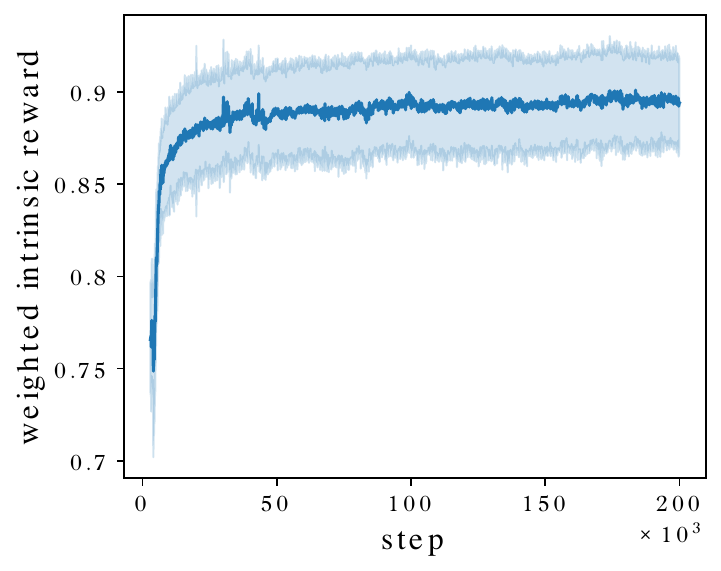}}\hfill
		\subcaptionbox{Average loss in credit assignment decreases over time.\label{ca_loss}}{\includegraphics[width=0.24\linewidth]{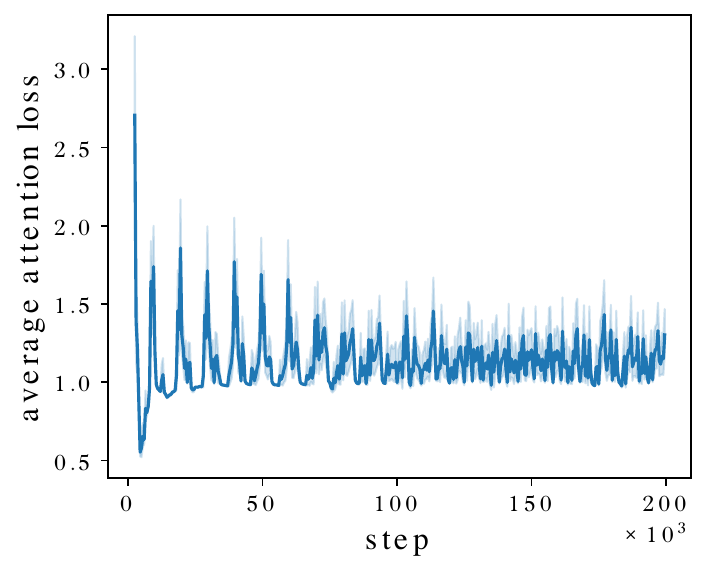}}\hfill
		\subcaptionbox{Average state prediction loss in curiosity decreases over time.\label{forward_loss}}{\includegraphics[width=0.24\linewidth]{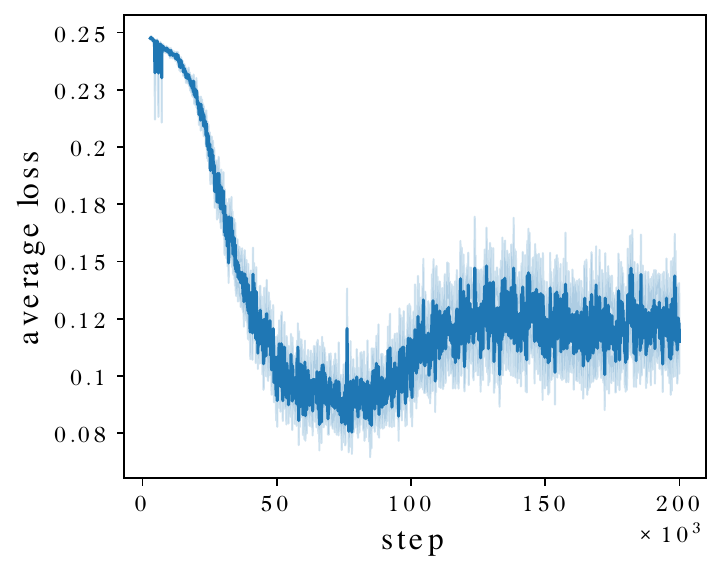}}\hfill
		\subcaptionbox{Average action prediction loss in curiosity decreases over time.\label{inv_loss}}{\includegraphics[width=0.24\linewidth]{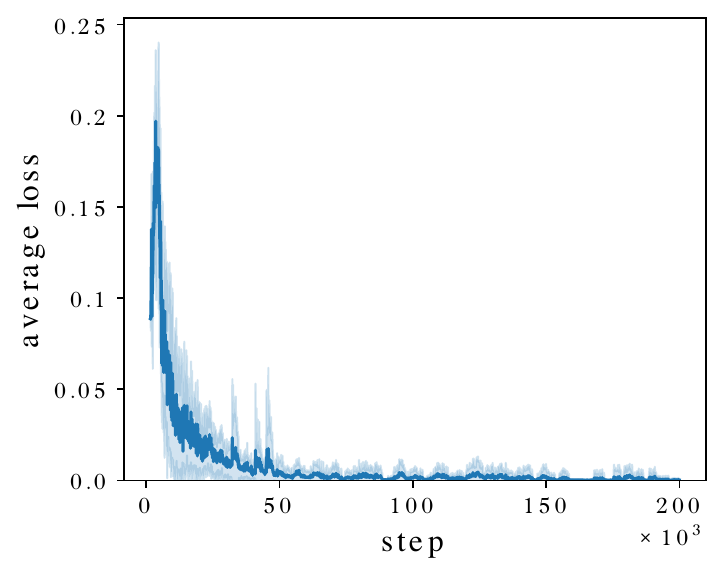}}\hfill
		\vspace*{-0.2cm}
		\caption{Training results}
		\label{trainingresults} 
	\end{figure*}

The two short-term rewards on auction utility and resource utilization are available immediately after the auction round. The two long-term rewards on offloading failure rate (OFR) and fairness are available after a $2000$-time-step delay. 

The same simulator is used for both offline training and online testing. However, there are significant differences to the environment setups. In the training evironment, besides bidders and the auctioneer, there is a coordinator agent that is only active during the outer-loop training phase to learn parameters for a generic model (Sec.~\ref{sec:federated}). The generic model is incrementally updated and used to initialize all local agents at the beginning of each inner-loop training phase. During every inner-loop, each bidder randomly selects a preference vector for the objectives and acts independently.

In the test environment, there is no coordinator agent, the bidders are initialized with the generic MOODY model in the beginning of the simulation, and they randomly select a preference vector at random intervals. Throughout evaluation, their credit assignment modules continuously predict rewards and trigger a short, adaptive retraining cycle according to Sec.~\ref{sec:adaptivecreditassignment2}. Besides these setup differences, the test environment also differs significantly from the training environment in resource capacity, vehicle arrival rate and speed, and traffic light phases. Table \ref{tab:setupdiff} summarizes the differences.

\subsection{Performance results}
\label{sec:eval}

All modules of \textbf{MOODY} converge to a local optimum in the training cycle (Fig.~\ref{trainingresults}). In the low-contention training setup, we reach close to optimal long-term objectives (i.e., OFR$\to 0$, fairness $\to 1$). 

In the following inference/retrain cycles, we compare the performance of \begin{inparaenum}[1)] \item \textbf{MOODY} bidders initialized with the generic model for multiple objectives, \item \textbf{DRACO2} bidders with the state-of-the-art single-objective algorithm from \cite{tan2022learning}, pretrained independently with a scalarized objective, \item \textbf{AC} bidders with only the actor-critic module. \end{inparaenum} The tests are run separately, each test has only one algorithm for all bidders in the simulation and run multiple times. We report the confidence intervals across all runs. In all tests, bidders' preference vectors change randomly over time, drawn from a uniform distribution. 

During testing, each MOODY bidder decides independently whether to trigger a retraining cycle. In our simulation, once retraining is triggered, the modules learn with $1$ shot in each retraining cycle. The DRACO2 bidders are retrained for a fixed $10$k time steps at the beginning of the deployment in the test environment. The AC bidders are not retrained. 

	\begin{figure*}[t]
		\centering
		\subcaptionbox{MOODY long-term individual OFR is $16\%$ lower than DRACO2, $30\%$ lower than AC.\label{separateTestofr}}{\includegraphics[width=0.24\linewidth]{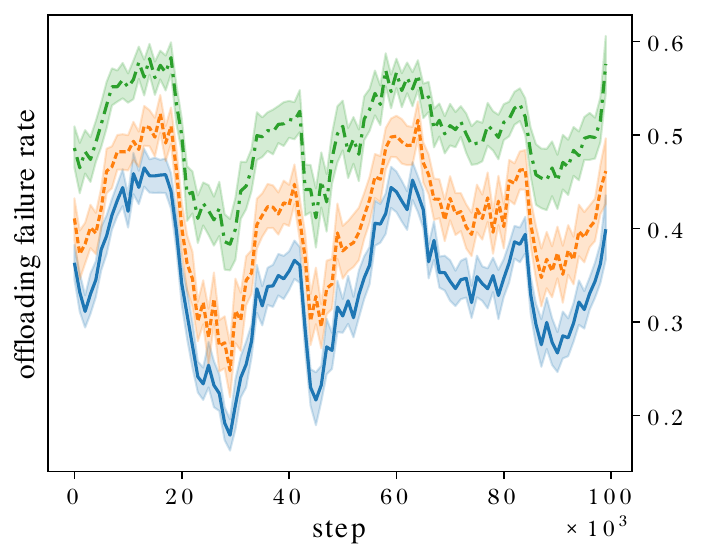}}\hfill
		\subcaptionbox{MOODY achieves average fairness of 0.92, compared to DRACO2: 0.89 and AC: 0.86.\label{separateTestfairness}}{\includegraphics[width=0.24\linewidth]{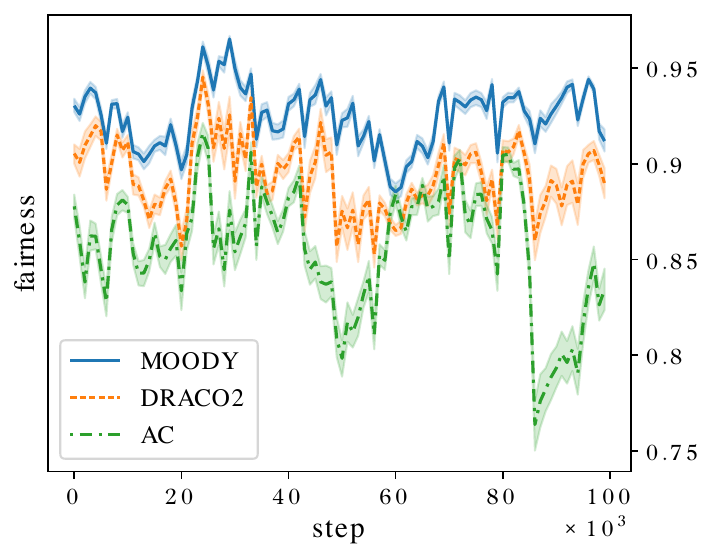}}\hfill
		\subcaptionbox{MOODY bidders' retrain cycles are $12\%$ of the time (gray lines are retrain cycles). \label{adaptivelearningresult}}{\includegraphics[width=0.24\linewidth]{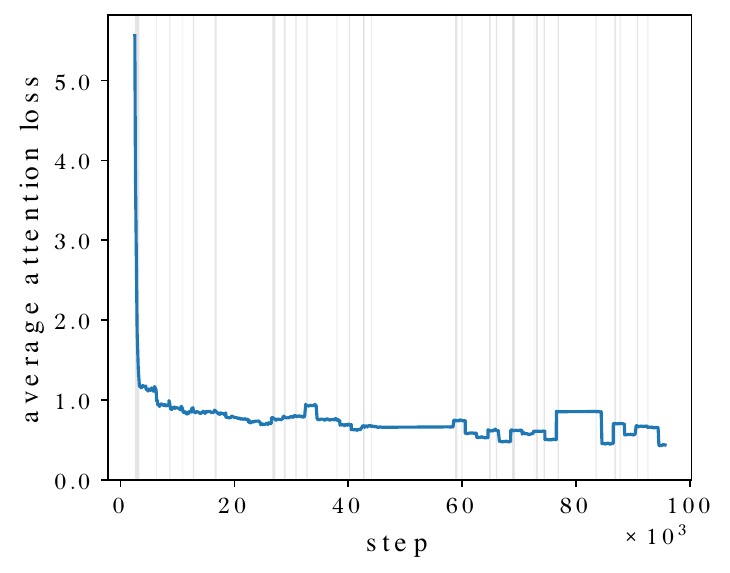}}\hfill
		\subcaptionbox{Achievement of high system fairness and low individual OFR is correlated.\label{fairnessVsOfr}}{\includegraphics[width=0.24\linewidth]{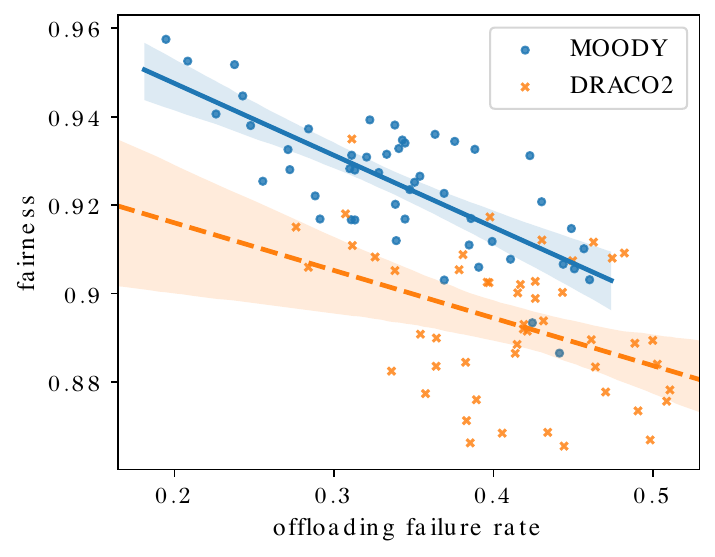}}\hfill
		\vspace*{-0.2cm}
		\caption{Objective achievement in test and retraining cycles}
		\label{testobjective} 
	\end{figure*}

Figures \ref{separateTestofr} and \ref{separateTestfairness} compare performance on the achievement of system long-term fairness and individual long-term OFR. In fact, in all four objectives, MOODY outperforms other bidders: MOODY's fairness score is close to $1$, compared to DRACO2's $0.89$ and AC's $0.86$; MOODY achieves $16-30\,\%$ lower offloading failure rate. Our results also show that MOODY achieves $46-77\,\%$ higher utility and $5-14\,\%$ less system utilization --- although the average utilization with MOODY and DRACO2 bidders are similar, MOODY lowers load variance by $19\,\%$ compared to DRACO2. With low variance, it is easier to plan long-term resource availability, saving cost while keeping the same service level.

	\begin{figure}[t]
		\centering
		\includegraphics[width=0.95\linewidth]{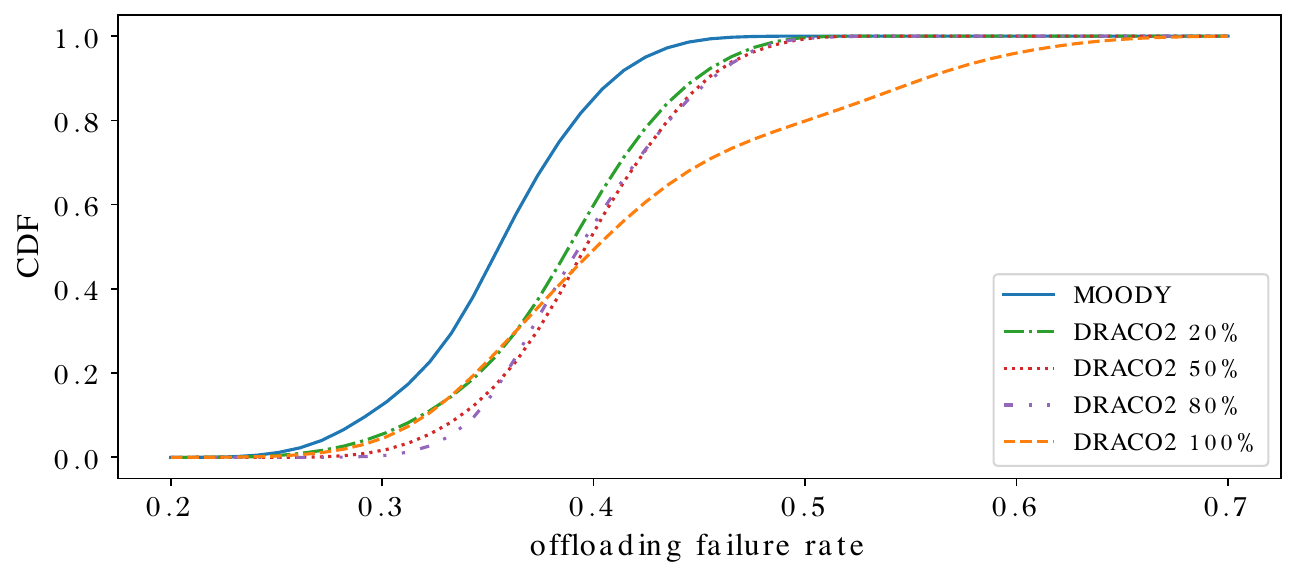}\hfill
		\vspace*{-0.2cm}
		\caption{In a heterogeneous test environment with competing algorithms, DRACO2 performance improves with MOODY unimpacted. System fairness also improves.}
		\label{heterogeneous} 
	\end{figure}

Fig. \ref{adaptivelearningresult} shows an example of how retraining contributes to the decrease in prediction loss for one of the bidders: as explained in Sec.~\ref{sec:adaptivecreditassignment2}, the retraining cycles are triggered by low reward prediction accuracy. The vertical gray lines are where retraining cycles occur. The bidder triggers a one-shot retraining cycle whenever the prediction accuracy of rewards reduces to below the moving average of the past $10$ prediction accuracies. As shown in Fig.~\ref{adaptivelearningresult}, the retraining cycles are frequently, almost continuously triggered in the beginning of the deployment in test environment. Overall, the MOODY bidders spend $9-15\,\%$ of time in retraining cycles.

Fig. \ref{fairnessVsOfr} shows correlation between achievements of the two long-term objectives: improvement in fairness is correlated to reduction in failure rate. In fact, we also see such correlation between other objectives. The reward signals on the system objectives help bidders learn this correlation, and by considering system objectives, the bidders effectively earn higher reward on their individual objectives, at the same time the auctioneer and commodity sellers achieve their objectives through incentivization.

All of the evaluations in Fig.~\ref{testobjective} are done with the same type of algorithms in the test environment (i.e., ``homogeneous''). However, in real life, vehicle users may run different algorithms (i.e., ``heterogeneous''). In Fig.~\ref{heterogeneous}, we show the cumulative distribution function (CDF) of each bidder's OFR performance, when the two algorithms compete in the same environment. 

The blue solid line labeled ``MOODY'' shows the performance of MOODY bidders in either the homogeneous (all-MOODY) or the heterogeneous environments with different percentage of MOODY bidders--their average performance in all environments are hardly different. In other words, they are unimpacted by the existence of other algorithms. Hence, to simplify the figure, we show their performance in one single curve. 
The dotted lines show performance in environments with different mix of MOODY and DRACO2 bidders. The rightmost line shows average performance of DRACO2 bidders in a homogeneous, all-DRACO2 environment, which has the worst performance of all environments.
Interestingly, in all of the heterogeneous environments, DRACO2 bidders' OFR performance improved, compared to the all-DRACO2 environment, reducing the difference to MOODY bidders by 50\%. Overall system fairness also improved significantly. These improvements do not depend on the percentage of MOODY bidders in the environment, indicating that even the presence of very few MOODY bidders can enhance overall performance. 

To summarize: we test MOODY's transfer learning capability by evaluating its performance in more dynamic test environments and allowing the bidders to change their objective preferences. Evaluation results show that \begin{inparaenum}[1)] \item bidders initialized with MOODY and adaptively retrained outperform bidders with other state-of-the-art learning algorithms in all objectives; \item the MOODY bidders demonstrate good generalization and transfer learning property, adapting to preference changes and dynamicity in the environment; \item the presence of MOODY bidders in the environment improves the performance of bidders with other algorithms and system overall fairness. \end{inparaenum}

\subsection{Sensitivity analysis}
\label{sec:sensitivity}

	\begin{figure}[t]
		\centering
		\subcaptionbox{Individual OFR \label{sensitivityofr}}{\includegraphics[width=0.48\linewidth]{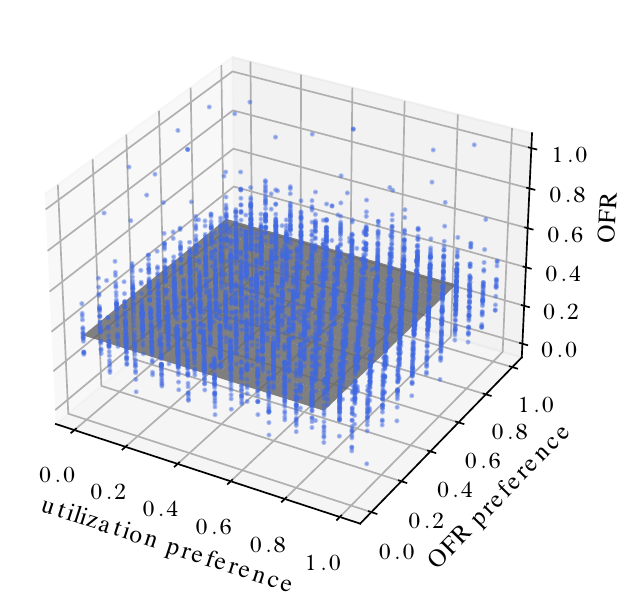}}\hfill
		\subcaptionbox{System fairnness\label{sensitivityfairness}}{\includegraphics[width=0.48\linewidth]{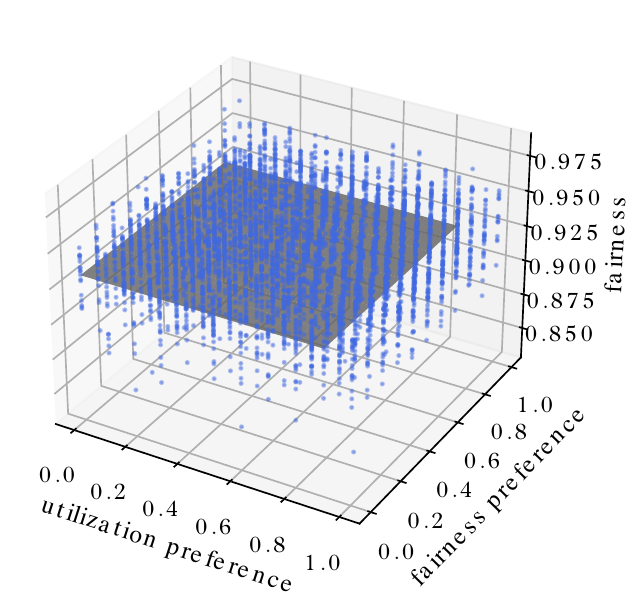}}\hfill
		\vspace*{-0.2cm}
		\caption{Sensitivity analysis shows target achievement is not sensitive to changes in preference.}
		\label{sensitivityPreference} 
	\end{figure}

First, we test the sensitivity of reward achievement to changing user preferences, based on data collected in inference/retraining cycle. We show in Fig.~\ref{sensitivityPreference} the sensitivity of the two long term objectives OFR and fairness to user preferences of OFR, fairness and resource utilization. As expected (see Sec.~\ref{sec:solution}), the rewards are not sensitive to different preference vectors. The user-specific optimal solutions are close to the initial generic model, needing only a few shots of retraining.

\begin{figure}[t]
		\centering
		\subcaptionbox{Individual OFR \label{sensitivityofr}}{\includegraphics[width=0.48\linewidth]{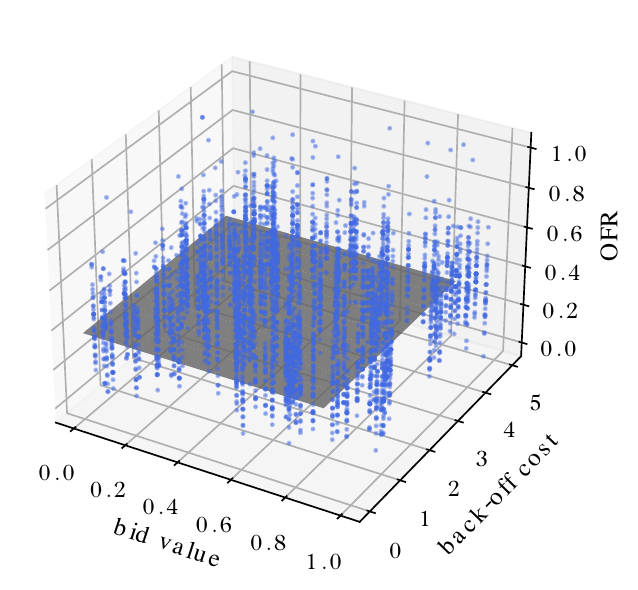}}\hfill
		\subcaptionbox{System fairness\label{sensitivityfairness}}{\includegraphics[width=0.48\linewidth]{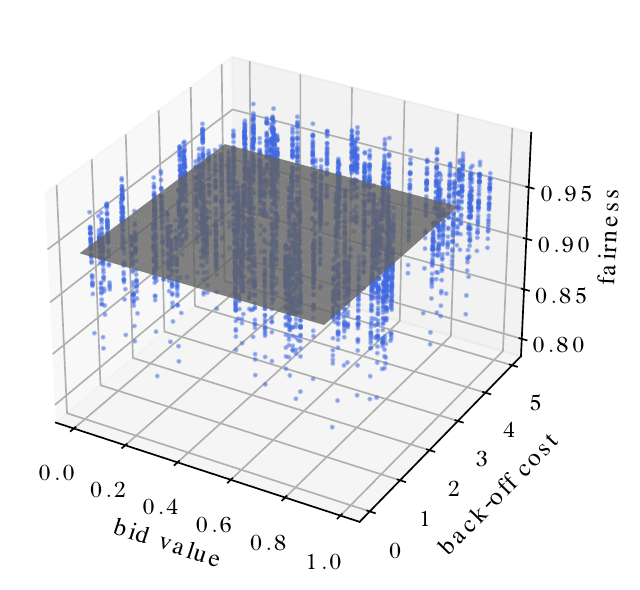}}\hfill
		\vspace*{-0.2cm}
		\caption{Sensitivity analysis shows target achievement is not sensitive to changes in hyperparameters.}
		\label{sensitivity} 
	\end{figure}

Next, we test the sensitivity of our solution to different hyperparameter inputs. We give two hyperparameters to each MOODY bidder at the time of initialization: \begin{inparaenum}[1)] \item the bid valuation $v_{m,k}$ that is private to each bidder $m$, and specific to each service request type $k$, and \item backoff cost $q_{m,k}$ that is private to each bidder and reciprocal to the time-to-deadline (Sec.~\ref{sec:utility}). \end{inparaenum} We change the value of these hyperparameters and show in Fig.~\ref{sensitivity} that MOODY is a robust algorithm that is insensitive to hyperparameter changes. 

\section{Practicality considerations}
\label{sec:chiptest}

\begin{table*}[t]
 \fontsize{8}{10}\selectfont
 \centering
 \begin{minipage}{\textwidth}
 \captionof{table}{Performance test}
 \label{tab:perfomancetest}
 \begin{tabular}{ >{\centering}p{0.17\textwidth}  >{\centering}p{0.20\textwidth}  
 >{\centering}p{0.14\textwidth}  >{\centering}p{0.20\textwidth}  
 >{\centering\arraybackslash}M{0.14\textwidth} }
 Modules & \multicolumn{2}{c}{Training} & \multicolumn{2}{c}{Inference}\\
 \cline{2-5}
    & Nr.calls & Per call (millisec)& Nr.calls & Per call (millisec)\\
 \toprule
 RL+credit & 108 & 5484 & 431 & 29\\
 supervised & 108 & 112 & 431 & 0\\
 curiosity & 197 & 3092 & 431 & 0\\
 data prep & 1275 & 10 & 431 & 29\\
\midrule
 \makecell{\textbf{Time per shot}\\(tested with Nano)} & max. of modules+data prep & \textbf{5494} &  sum of modules+data prep &\textbf{58}\\
\makecell{\textbf{Time per shot}\\(estim. with AGX Orin)} & $1/10^{th}$ of Nano & \textbf{550} & $1/10^{th}$ of Nano &\textbf{6}\\
\bottomrule
 \end{tabular}
\end{minipage}
\end{table*}

To speed up training, we train each module asynchronously in federated learning (Sec.~\ref{sec:federated}). Before each inner loop begins, the local agent is initialized with generic model parameters. Then, the agent joins the auction whenever it receives a request. Since each local agent receives requests randomly and makes independent decisions, they finish the inner-loop training phase at different time steps. Furthermore, each module trains at different time intervals. Asynchronous training reduces peak data rate (i.e.\,  the maximum data volume transmitted over a network per second). It also reduces training time: the time for training once (i.e., one-shot) is the \textbf{maximum} duration among the modules \begin{inparaenum}[1)] \item RL with credit assignment (\emph{RL+credit}), \item supervised learning (\emph{supervised}) and \item curiosity learning (\emph{curiosity})\end{inparaenum}. Without the asynchronous training, one-shot training time would be the \textbf{sum} of all modules. More importantly, asynchronous training reduces the online retraining time after deployment.

After deployment, bidders decide independently when to retrain the model to adapt to new objectives and environments (Sec.~\ref{sec:adaptivecreditassignment2}). The retraining is easily separated from the main program that infers bidding decisions in runtime (i.e., \emph{out-of-critical-path}). With retraining off the critical path, we can ensure fast decision-making even with retraining.

We test real-life training, retraining and inference speed of our algorithm on an Nvidia Jetson Nano single-board computer with GPU. The single-board computer simulates an onboard unit of a vehicle, or a bidder in the auction. We run the training and inference repeatedly and record the average time for one shot. The results are shown in Table~\ref{tab:perfomancetest}. Besides data preparation (e.g.\, input data formatting, reshaping, stacking, etc. that is done once for all modules), time for one-shot training is the maximum time among all modules, and time for inference is the sum of all modules. We provide the measured performance on Nano, and an estimated performance on the newer AGX Orin. Although the theoretical performance difference between the two is $>100$ times, multiple benchmark tests on various AI applications show a more realistic performance difference of ca. $10$ times (see Nvidia website for Jetson modules technical specifications and benchmarks). We therefore estimate that with AGX Orin, training one shot takes ca. $550$ms, and inference takes ca. $6$ms. Speed can be further increased through fewer layers and nodes, smaller batch size, shorter input length, etc. The analysis of performance impact is left to future work.

These results show that despite the complexity of the proposed solution, bidders can perform runtime inference, on current hardware, with a reaction time of $6$ms. V2X applications (e.g., segmentation, motion planning) typically run on the time scale of seconds, an inference speed in milliseconds makes our model a good candidate for real-life deployment. The retraining cycle is longer, for which we believe that \emph{out-of-critical-path} few-shot retraining holds great promise. Even without that optimization, the retraining cycle lasts only a few seconds, well below the frequency of changes in a V2X environment that may trigger retraining.

 \section{Conclusion \& future work}
\label{sec:conclusion}

We combine offline federated learning and online few-shot learning to solve an MOP in a dynamic environment. Through extensive offline training, we get an optimal initial model that learns the best initialization point. From this point, it can quickly find a solution on the Pareto frontier, even without retraining, when the agent's objectives change. Only in a significantly different environment, we allow each bidding agent adaptive, online, few-shot retraining to customize its model, needing very few data points. 

We show empirically that our new multi-objective algorithm outperforms the benchmark algorithms in all objectives. Furthermore, our algorithm increases bottom-line resource efficiency, such that other algorithms in the environment also benefit from improved offloading success rate and fairness.

Our algorithm can be easily modularized, each module trained separately and asynchronously. Coupled with the adaptive few-shot online training method, the algorithm is a very good candidate for real-life deployment. 

Currently, we simulate agents' preference of objectives with uniform randomly generated weights, and scalarize the rewards with a linear objective function, with the assumption that the individual objectives are independent from each other. There are two potential improvements to this approach:\begin{inparaenum}[1)] \item the method for sampling preferences may impact the approximation of the Pareto frontier and the performance of the initial model. Future work should consider different sampling methods such as proposed in~\cite{ryu2009pareto} and \cite{khorram2014numerical}. \item Simulation results show that agents learn the correlation between different objectives. In fact, multiple objectives in real-life are typically correlated to each other. There may exist a hierarchy or network of objectives, and we should guide the learning process with this knowledge of the objective structure in our future work.\end{inparaenum}

\bibliographystyle{IEEEtran}

\end{document}